\documentclass[11pt,letterpaper]{article}
\usepackage{cogsys}
\usepackage[T1]{fontenc}
\usepackage{times}
\usepackage[pdftex]{graphicx} % use this when importing PDF files

\usepackage{natbib}
\cogsysheading{11}{2021}{1--**}{9/2021}{11/2021}
\ShortHeadings{Preliminary Work on a Motif Detector}
              {Yarlott \it{et al.}}

\begin{document} 
\title{Finding Trolls Under Bridges: Preliminary Work on a Motif Detector}

\author{W. Victor H. Yarlott$^1$}{wvyar@cs.fiu.edu}
\author{Armando Ochoa$^1$}{aocho032@fiu.edu}
\author{Anurag Acharya$^1$}{aacha007@fiu.edu}
\author{Laurel Bobrow$^2$}{lbobrow@sift.net}
\author{Diego Castro Estrada$^1$}{dcast230@fiu.edu}
\author{Diana Gomez$^1$}{dgome133@fiu.edu}
\author{Joan Zheng$^2$}{jzheng@sift.net}
\author{David McDonald$^2$}{dmcdonald@sift.net}
\author{Chris Miller$^2$}{cmiller@sift.net}
\author{Mark A. Finlayson$^1$}{markaf@fiu.edu}
\address{$^1$Knight Foundation School of Computing and Information Sciences, Florida International University, 11200 SW 8th St., Miami, FL 33199 USA}
\address{$^2$Smart Information Flow Technologies (SIFT), 114 Waltham Street, Ste. 24, Lexington, MA 02421 USA}

\vskip 0.2in
 
\begin{abstract}
	Motifs are distinctive recurring elements found in folklore that have significance as communicative devices in news, literature, press releases, and propaganda. Motifs concisely imply a large constellation of culturally-relevant information, and their broad usage  suggests their cognitive importance as touchstones of cultural knowledge, making their detection a worthy step toward culturally-aware natural language processing tasks. Until now, folklorists and others interested in motifs have only extracted motifs from narratives manually. We present a preliminary report on the development of a system for automatically detecting motifs. We briefly describe an annotation effort to produce data for training motif detection, which is on-going. We describe our in-progress architecture in detail, which aims to capture, in part, how people determine whether or not a motif candidate is being used in a motific way. This description includes a test of an off-the-shelf metaphor detector as a feature for motif detection, which achieves a $F_1$ of 0.35 on motifs and a macro-average $F_1$ of 0.21 across four categories which we assign to motif candidates. 
\end{abstract}

\section{Motifs as a Source of Cultural Information}
\label{sec:intro}

Motifs are distinct, recurring narrative elements found in folklore and, more generally, culturally inflected materials. Motifs are interesting because they provide a compact source of cultural information: many motifs concisely communicate a related constellation of cultural ideas, associations, and assumptions. For example, ``troll under a bridge'' is an example of a motif common in the Western cultures that have roots in Scandinavia. To those familiar with the motif, it entails a number of related ideas that are by no means directly communicated by the surface meaning of the words: the bridge is along the critical path of the hero, and he must cross to achieve his goal; the troll often lives under the bridge, crawling out to waylay innocent passers-by; the troll charges a toll or exacts some other payment for crossing the bridge; the troll is a squatter, not the `officially' sanctioned master of the bridge; the troll enforces his illegitimate claim through threat of physical violence; and the hero often ends up battling (and defeating) the troll instead of paying the toll.

Because of this density of information, motifs are often retained as a tale is passed between cultures and down generations, and folklorists have observed that a tale's specific composition of motifs can be used to trace the tale's lineage \cite[Part 4, Chapter V]{thompson1977folktale}. This has led folklorists to construct motif indices that identify motifs and note their presence in specific tales (usually as represented in a particular folkloristic collection).  The most well-known motif index is the Thompson motif index (TMI) by \cite{thompson1960motif}, which references tales from over 614 collections, indexed to 46,248 motifs and sub-motifs, 41,796 of which have references to tales or tale types. In Thompson's index each motif is given a designating code; for example, ``troll under a bridge'' is referenced by the codes G304 and G475.2. In this case, ``troll under a bridge'' is represented by two motifs as Thompson generalizes trolls to ogres, a general class of monstrous beings; thus, the motifs are ``troll as ogre'' (G304) and ``ogre attacks intruders on bridge'' (G475.2).

Thompson informally defines a motif as items ``worthy of note because of something out of the ordinary, something of sufficiently striking character to become a part of tradition, oral or literary. Commonplace experiences, such as eating and sleeping, are not traditional in this sense. But they may become so by having attached to them something remarkable or worthy of remembering'' \cite[p. 19]{thompson1960motif}. He notes that motifs generally fall into one of three subcategories: an event, a character, or a prop \cite[pp. 415--416]{thompson1977folktale}. Here we give an example of each (with their associated Thompson's motif code):

A \textbf{hero rescuing a princess} (B11.11.4) is perhaps one of the most well-known  event motifs in western culture.  Ask a westerner the following question: ``A princess has been kidnapped: who kidnapped her, who rescues her, and what does the rescuer need to do to effect a rescue?'', and common answers will be ``a dragon kidnapped her, the knight must rescue her, and he must kill the dragon.'' This motif may be the climax of the story, with a ``happily ever after'' ending just after the hero defeats the dragon, or it may just happen in the course of a story: in \textit{Ivan Dogson and the White Polyanin}, a Russian folktale \cite[Tale \#139]{afanasev1957narodnye}, Ivan slays three dragons, each with more heads than the last, rescuing a princess each time. The motif is prolific, found across the tales, literature, and movies of many cultures.

\textbf{Old Man Coyote} (A177.1) is a character motif: known in some Native American Indian tribes as Coyote, he is one of the most recognizable gods. In Native American Crow folklore, Old Man Coyote creates the earth and all the creatures on earth. He travels the world, teaching the animals how they should behave. Old Man Coyote, however, is far from a noble and elegant creator. He creates ridiculous costumes and tries to trick the Crow tribe into wearing them, only to be run off. He purposefully bungles rituals to produce food, such as transforming skin from his back to meat, in order to guilt his guests into performing the ritual correctly to get free food, later performing it correctly to discredit his former guests when they tell others he erred. Anywhere Old Man Coyote is referenced, he calls to mind someone who has done great things, but is lazy and often far too clever for their own good, falling pray to their own cunning.

A \textbf{magic carpet} (D1155) is a prop that allows the hero to fly through the sky, and is familiar to anyone who has watched Disney's \textit{Aladdin}. In \textit{One 	Thousand and One Nights}, Prince Husain encounters a merchant selling a carpet for an outrageous price; the merchant says:  ``O my lord, thinkest thou I price this carpet at too high a value? \ldots Whoever sitteth on this carpet and willeth in thought to be taken up and set down upon other site will, in the twinkling of an eye, be borne thither, be that place nearhand or distant many a day's journey and difficult to reach'' \cite[p. 496]{burton2009nights}. Solomon, said to be the third king of Israel, was said to have a carpet 60 miles on each side that could transport him vast distances in a short amount of time. In Russian hero tales, magic carpets are common items that aid the hero in his quest.

While the above examples are drawn from folklore, motifs have importance beyond folktales: they occur in everyday speech, news stories, press releases, propaganda, novels, movies, plays, and anywhere that cultural materials are found. A powerful modern example is the use of the \textit{Pharaoh} motif in modern middle eastern discourse. The Pharaoh appears in Qur'an, and comes into conflict with Moses and his attempts to free the Hebrews from Egyptian slavery.  The Pharaoh is an arrogant and obstinate tyrant who defies the will of God and is punished for it. In modern Islamist extremist narratives, the Pharaoh is a symbol of struggles against anti-Islamic regimes and has been invoked against leaders such as Anwar Sadat of Egypt, Ariel Sharon of Israel, and George W. Bush, whom Osama bin Laden referred to as the ``pharaoh of the century'' as noted by \cite{halverson2011master}.

While motifs appear in many places, it is not always clear when they are being used to communicate the greater constellation of their folkloric meaning: the raw text ``pharaoh'' itself could be an allusion to historical tyranny but it could easily appear in a document describing the leaders of Ancient Egypt. It is therefore also necessary to be able to clearly distinguish between different usages of specific instances of motif text---what we refer to as ``motif candidates.'' We do so by assigning them to one of four categories: motific, referential, eponymic, or unrelated. We describe these four categories in more detail in section \S\ref{sec:procedure}.

We find that, in practice, motifs somewhat resemble figurative usage of language; however, it's not clear-cut. Calling an athlete Finn McCool may very well be their name in Ireland---or it might be alluding to the mythical hero Finn McCool to praise the athlete's physical prowess. Therefore, we believe that some degree of overlap between metaphor, simile, and motif exists, but that metaphor and simile detectors will not be off-the-shelf solutions for motif detection.

Because of their prominence and ubiquity, the ability to automatically detect motifs would open up a vast repository of important cultural information---something that as of yet, is unaddressed~\citep{acharya2021atlas}---to computational analysis, allowing systems to be built that leverage this powerful facet of human communication. Currently, motifs must be extracted by hand by trained cultural experts.

In this work, we outline preliminary efforts to build a system for automated motif detection. We briefly describe our ongoing annotation study (\S\ref{sec:data}), describe our on-going work in developing a pipeline based on how humans understand stories for the automatic detection of motifs (\S\ref{sec:architecture}), and present  preliminary results exploring metaphoricity as a feature for motif detection (\S\ref{sec:metaphor}). We also review related work to provide context (\S\ref{sec:related}). We conclude by describing potential future work and summarizing our contributions (\S\ref{sec:contributions}).

\section{Data Production}
\label{sec:data}

As a first step toward developing an automated system, we have undertaken an annotation study to produce data categorizing motifs as either \textit{motific}, \textit{referential}, \textit{eponymic}, or \textit{unrelated}. This on-going annotation will result in identifying and labeling approximately 21,000 motif candidates, for which analysis is still underway. We briefly describe the annotation study. This data will be used to validate our automatic system for detecting motifs.

\subsection{Selection of Motifs}
\label{sec:selection}
To begin selecting motifs, we needed to identify the cultural groups from which to draw motifs and their source folklore. This was done by checking two criteria: first, that there was a strong authoritative source of motifs (a motif index or folklore collection)---this made it easier for us to identify motif candidates. Second, a sizable population that was easily accessible---this was partially in service of an ongoing survey study as part of the same effort this work falls under, but also to ensure that we would easily be able to find annotators (although world events quickly made these concerns obsolete). Using these criteria, we selected Irish, Jewish, and Puerto Rican as the cultural groups to draw motifs from, as well as to recruit annotation candidates from.

We had several selection criteria: (1) motifs must be highly-referenced in social media or news, demonstrating they were in common use; (2) motifs must have a definitive source within the folklore of the cultural group, giving proof of relevance; and, (3) the motifs must have a high potential strength, which we subjectively estimated based on input from members of the cultural group and the frequency of usage of the motifs in social media and news.

\subsubsection{Examples}
We identified 22 character motifs, 11 prop motifs, and 6 event motifs for use in the study, of which 13 were Irish, 10 were Jewish, and 16 were Puerto Rican. Here, we provide sample motific usages:\\

Finn McCool, Irish:
\begin{quote}
     In fact, that was where he acquired the nickname, playing for the parish in a Co Kilkenny semi-final, a young \textbf{Finn McCool} with limbs that might bestride a bullock.
\end{quote}

Tower of Babel, Jewish:
\begin{quote}
    The first Slavic pope, John Paul II, swept joyfully through the streets of the Czechoslovak capital yesterday, proclaiming a unified Europe was at hand and communism had crumbled like the \textbf{Tower of Babel}.
\end{quote}

Coqui, Puerto Rican:
\begin{quote}
    We listen here for the first strains of the exotic, in the parrot's squawk, the \textbf{coqui's} clamor, the reggaeton beat - but also for the beleaguered howl of the native: the panther's scream, the ivorybill's rap, the Miccosukee's lilting tongue.
\end{quote}

\subsection{Selection of Data}
\label{sec:data-selection}
Selection of data was done using keyword searches of phrases related to motifs of interest, identified as described in \S\ref{sec:selection}, through NexisUni, a university-focused version of LexisNexis, a product that makes news articles easily searchable. The resulting articles of this basic keyword search were downloaded and converted from rich-text format to plain-text format. This data consists of news articles that LexisNexis identifies to have text matches for phrases relevant to the motif (e.g., ``Finn McCool'').

These plain-text news articles were then further processed by our own Lucene-based lexical matcher, which uses a hand-crafted ruleset to identify the surface form of known motifs; these surface forms are not necessarily true motific usage (e.g., it could be an article on the origin of ``Amalek'')---thus, we refer to them as motif \textit{candidates}. We use these candidates to verify the presence of motifs and tag the spans of text containing them in a stand-off annotation. The articles are then chosen from randomly to avoid any temporal or geographic dependencies that might be order dependent, and the articles and annotations are then used for both the annotation (using the brat rapid annotation tool \citep{brat}) and as input to the pipeline.

\subsection{Annotator Training}
\label{sec:training}
Annotation was done by three pairs of annotators hired for their membership in the cultural groups (Irish, Jewish, and Puerto Rican) that were chosen for the accessibility of their motifs and their proximity to our researchers, as described in \S\ref{sec:selection}. Each pair consisted of annotators from a single one of these three cultural groups, tasked with annotating Irish, Jewish, or Puerto Rican motifs as appropriate. All annotators had or were in the process of obtaining college degrees.

Annotators were given a single two-hour session of training that included reviewing the annotation guide, covering any questions or concerns, and running through a small sample annotation together as a team. Further, a two-hour adjudication session was held every week to cover the week's annotations: while these served to help reinforce the annotator's skills, they also served to produce adjudicated data for machine learning. Annotators were not allowed to communicate with each other regarding the annotations until these adjudication meetings.

Annotators are also given access to an annotation guide that provides a high-level description of annotations, descriptions of what to annotate, tutorials of how to use the tool, and detailed descriptions of the motifs drawn from discussion with in-culture members. Annotators were allowed to consult the annotation guide at any point during annotation.

\subsection{Annotation Procedure}
\label{sec:procedure}
Annotators used the brat rapid annotation tool with pre-marked text spans \citep{brat}. This was displayed as a text file with several snippets of articles containing highlighted spans representing the motif candidate that the annotators would have to change to the correct annotation. Each span represented a motif candidate with its extent drawn from our lexical matcher and it's label simply being the motif: e.g., \texttt{finn-mccool}, \texttt{amalek}, etc. Annotators were instructed to tag each of these as one of four categories:

\begin{description}
	\item[Motific] A usage that intends to invoke the cultural associations of a motif (e.g. referring to something large and monstrous as a ``behemoth'').
	\item[Referential] A usage that directly refers to the folklore origin of the motif or its definition (e.g., discussing the origin of the term ``behemoth'').
	\item[Eponymic] A usage that references the motif in a name---this distinction is made because while it is highly similar to motific usage, it is typically not used as such (e.g., the band ``behemoth'' may be referred to with no additional meaning beyond the band).
	\item[Unrelated] A usage that is unrelated to the cultural group or cannot be established as directly related (e.g., ``behemoth'' as a monster in a game).
\end{description}

Once annotators completed a batch of annotation (a process that took roughly a week and produced anywhere from 500-1000 candidates, depending on the group and the selection of articles), an adjudication meeting was held. Inconsistent tags were reconciled via discussion and agreement on the correct tagging; this serves not only to increase the amount of gold-standard data, but also to reinforce annotator training and address any issues that arose during the annotation process.

\subsection{Preliminary Agreement}
\label{sec:agrement}
For this annotation study, we expect annotator agreement to remain above 0.55 Fleiss' kappa: we show preliminary results for the first two batches after training until reaching the desired Fleiss' kappa in Table~\ref{tab:annotation_results}. These results are preliminary, but suggest that humans with a strong background in the relevant culture can reliably distinguish motific, eponymic, referential, and unrelated candidates of a motif. Table~\ref{tab:annotation_results} also shows preliminary tag counts and number of candidates annotated with a high level of agreement---this distribution is in line with our expectations that motifs are relatively rare in clear motific usage as opposed to eponymic or referential usage, based on early exploratory work and the understanding that using motifs in a motific way requires a deeper level of understanding of the motif.

\begin{table}[t]
	\vskip -0.15in
	\caption{Preliminary agreement and counts from our in-progress annotation study.}
	\label{tab:annotation_results}
	\begin{small}
		\begin{center}
			% \vskip -0.10in
			\begin{tabular}{l|r|r|r|r|r|r}
				\hline
				\abovespace\belowspace
				Group & Agreement ($F_\kappa$) & \# Motific & \# Referential & \# Eponymic & \# Unrelated & Batch Size \\
				\hline\abovespace
				Irish        & 0.559 &  11 & 292 & 353 & 207 & 863 \\
				Irish        & 0.699 &  46 & 355 & 323 & 254 & 978 \\
				Jewish       & 0.579 &  56 & 232 & 246 &  18 & 552 \\
				Jewish       & 0.638 & 133 & 374 & 348 &  56 & 911 \\
				Puerto Rican & 0.552 &  66 & 427 & 237 & 134 & 864 \\
				Puerto Rican & 0.680 &  19 & 433 & 212 & 174 & 838 \\
				\hline
				\multicolumn{2}{l|}{Total} & 331 & 2,113 & 1,719 & 843 & 5,006 \\
				\hline 
			\end{tabular}
		\end{center}
		\vskip -0.10in
	\end{small}
\end{table}

\section{Architecture}
\label{sec:architecture}
\begin{figure}[t]
	\vskip 0.05in
	\begin{center}
		% \setlength{\epsfxsize}{3.5in}
		% \centerline{\epsfbox{cascade.eps}}
		\includegraphics[width=\textwidth]{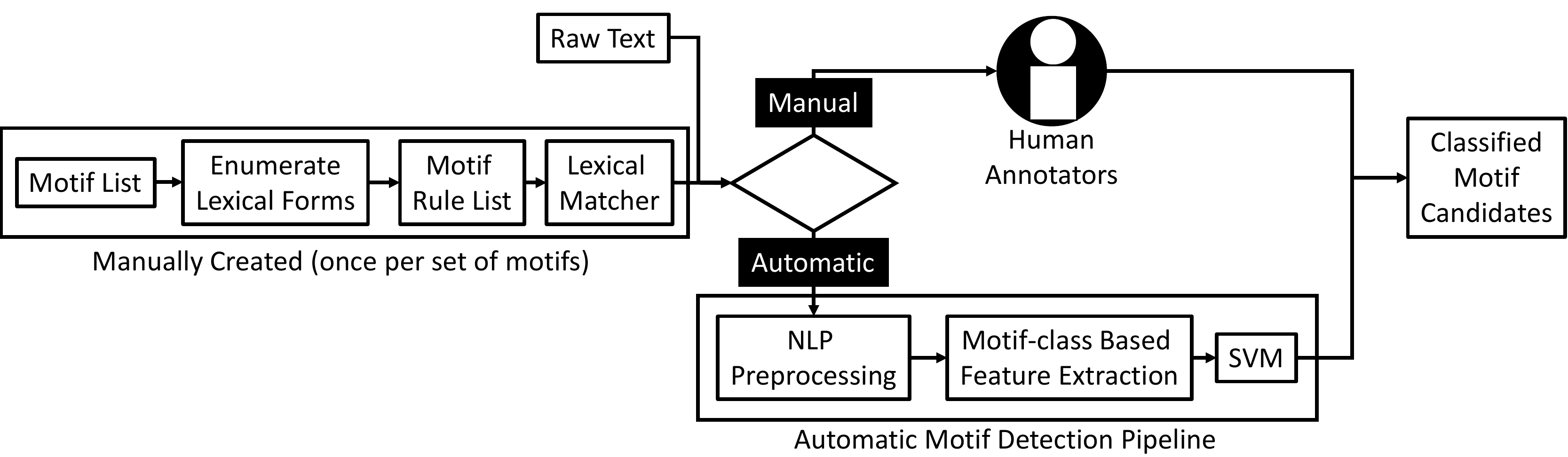}
		% \vskip 0.1in
		\caption{End-to-end Motif Detection Pipeline. Displayed are necessary manual preprocessing steps, and our automatic system, which we are validating with annotated data, performs the task assigned to our human annotators: identifying which of the candidates from our high-recall lexical matcher are motific, eponymic, referential, or unrelated.} 
		\label{fig:architecture}
	\end{center}
	\vskip -0.2in
\end{figure}

We show our end-to-end pipeline in Figure~\ref{fig:architecture}. The majority of our automatic system is contained within the black square and will perform the same task as the human annotators with the same input: the raw article text with the spans containing candidate matches using a manually-created motif rule list.

The automatic system is designed as a traditional NLP pipeline (as opposed to a neural system) that caches the results of many different NLP tools, the output of which are parsed into features inspired by our understanding of humans read and understand stories, as well as intuitive constructs based on these ideas that target the three specific categories of motifs described by Thompson: characters (``the actors in a tale''), props (``certain items in the background of the action''), and events (``single incidents'') \cite[pp. 415--416]{thompson1977folktale}. In \S\ref{ssec:features}, we discuss what features we are implementing to target these categories of motifs.

\subsection{Manual Motif List Creation}
Manual input is, at this point in time, a necessary evil---this step is required to identify a list of motifs of interest (as described in Section \S\ref{sec:selection}) and then to bootstrap initial lexical forms and write lexical matcher rules for these forms. These rules are used with a Lucene-based lexical matcher to identify the spans of text that are selected as motif candidates.

This procedure is done only once for each motif that is part of the list of motifs: once the lexical matcher's rules are set and run over the raw text, the stand-off brat annotation files are created, and both are ready to be passed on to either human annotators, if the work is being conducted manually as in our annotation study, or to our automatic motif detection pipeline. We imagine that in the future, this step could be automated through the development of an automatic parsing engine for motif indices, as described Yarlott \& Finlayson \citeyearpar[p. 7]{yarlott2016learning}.

\subsection{NLP Preprocessing}
\label{ssec:preprocessing}
For preprocessing, we use many off-the-shelf NLP modules, such as Stanford's CoreNLP \citep{stanfordCoreNLP} and OpenIE \citep{stanfordOpenIE}, and spaCy \citep{spacy}. Additionally, we use Sparser \citep{mcdonald1992efficient, mcdonaldlearning} to extract high-precision information related to the motif candidates.

We also use modules based on figurative language: our own reimplementations of the DeepMET metaphor detector \citep{su2020deepmet} and a neural simile detection system \citep{liu2018neural}. These modules are used, in part, to help catch the overlap between figurative language and motific usage, as well as to discriminate against categories that have a high degree of non-figurative usage, such as referential examples.

We additionally run the animacy and character systems developed by \cite{jahan2018new, jahan2020straightforward}, our own reimplementation of a possession detector \citep{chinnappa2018mining}, the CAEVO event detector \citep{chambers2014dense}, and the SRL included in Story Workbench \citep{finlayson2008collecting, finlayson2011story} and the AllenNLP SRL \citep{Gardner2017AllenNLP}.

The end result of this preprocessing is a large cache of stand-off files containing per-document dependency parses, POS tags, NER tags, OpenIE triplets, coreference chains with accompanying animacy and character labels, events, and SRL relations; additionally, the preprocessing produces per-candidate metaphor, simile, and possession tags. All of these stand-off files are either processed into feature vectors (in the per-candidate cases) or further computed into useful feature vectors (in the per-document cases).

\subsection{Feature Extraction}
\label{ssec:features}
We draw substantial inspiration for our features from the theoretical description of motifs as characters, props, and events by Stith Thompson, with groups of features targetting specific classes of motifs.

\textbf{Character motifs} are very clearly related to animacy and characters in stories and so, we calculate distance, coreference, and semantic relation or grammatical participation between the candidates and animate or character coreference chains. Additionally, characters very often have a proper name, for which we target with NER features.

\textbf{Prop motifs} are also checked for semantic, grammatical, or possession relation with animate or character entities, as in motifs they are often used in relation to characters. We further check general semantic role and part-of-speech to target prop motifs.

To target \textbf{event motifs}, we use CAEVO to determine the participation in, co-occurrence with, and distance from events: these different features are because an event motif might be used \textit{as} the event itself, but it may also be nearby as part of a comparison or as further description. We also expect the entities taking part in an \textbf{event motifs} to be animate.

The end result of this feature extraction are per-candidate vectors of features, among which are: whether or not they are animate, participation in animate chains, participation in events, semantic relation to animate entities, grammatical relation to animate entities, nearby NER tags, and whether or not they are metaphoric or involved in simile.

\subsection{Motif Detection}
The features being used for motif detection generally fall into one of three categories: (1) their normalized relative position to a motif candidate (windows, co-occurrence, token distance, and parse distance); (2) their semantic relation with the motif candidate (SRL, IE, possession, discourse relation, and word sense); or (3) the usage of figurative language in relation to the motif candidate (simile and metaphor). Additionally, we are developing a new metaphor detector in hopes of better capturing where metaphoric and motific usage overlaps. Our detector also attempts to predict the motif type from the features and compare that to the actual, expected motif type based on the lexical form (e.g., ``Finn McCool'' is a character, not a prop). These features will then be fed into a means of classifying them as motific, referential, eponymic, or unrelated.

\section{Preliminary Results: Metaphor}
\label{sec:metaphor}
As a preliminary test, we reimplemented the DeepMET metaphor detector described in \cite{su2020deepmet} to determine if a motif candidate is in the same sentence as a metaphor. We use this test as a sanity check---intuitively, we expect motifs to primarily be figurative in use, as referring to someone as ``Amalek''---an antagonist in Jewish stories---or comparing something to a ``coqui''---a frog native to Puerto Rico that features prominently as its mascot---are not literally stating that they are a heinous ruler from millennia past or a frog. Interestingly, however, we find that this is not exactly the case: in Table~\ref{tab:results}, we present the results of testing the metaphor-based feature.

\begin{table}[t]
	\vskip -0.15in
	\caption{Results of using the DeepMET metaphor detector by \cite{su2020deepmet} to detect motifs. These results were obtained using part of the pipeline architecture under development as described in Section \S\ref{sec:architecture}.}
	\label{tab:results}
	\begin{small}
		\begin{center}
			% \vskip -0.10in
			\begin{tabular}{l|c|c|c|c|c}
				\hline
				\abovespace\belowspace
				Model & Motif $F_1$ & Eponym $F_1$ & Referential $F_1$ & Unrelated $F_1$ & Macro $F_1$ \\
				\hline\abovespace
				Metaphor Alone, SVM & 0.35 & 0.00 & 0.59 & 0.00 & 0.21 \\
				\hline 
			\end{tabular}
		\end{center}
		\vskip -0.10in
	\end{small}
\end{table}

The results demonstrate that while there is some overlap between metaphor and motific usage, it does not achieve substantial performance ($F_1 = 0.35$), with the feature performing better at detecting referential usage of motif candidates ($F_1 = 0.59$).

This result is not wholly unexpected. As we discussed briefly in \S\ref{sec:intro}, the relation between metaphor and motif is not 1-to-1, thus resulting in relatively low performance. We theorize that the relatively high performance at detecting referential usage may be due to referential usage, which we describe as either strict discussion of the folklore origin or a definition of the motif, are much more likely to use literal language. Therefore, the a \textit{False} metaphor tag may be an indicator of referential usage.

These results, which show that metaphor detection does not work as a drop-in motif detector, reinforce our belief that deeper systems based on story understanding and theory (as the origin of motifs \textit{is} stories) are necessary to detect motifs.

\section{Related Work}
\label{sec:related}
On the folklore side, there are many motif indices, with the Aarne-Thompson Motif-index of Folk-literature \citep{thompson1960motif} being the primary resource.  There are numerous other indices, most primarily focused on a specific culture. Thompson also has substantial discussion on motifs and the compilation of indices in his book \textit{The	Folktale}~\citep{thompson1977folktale}. While Thompson's motif index is perhaps the primary source of motif information used today, it has been criticized because of overlapping motif subcategories, censorship (primarily of obscenity), and missing motifs \citep{dundes1997motif}. These motif indices provide a substantial base for us to build upon and we draw heavily from both the Aarne-Thompson index as well as Tom Peete Cross' Motif-index of of Early Irish Literature~\citep{cross1952motif} and Dov Noy's Motif-index of Talmudic-Midrashic literature~\citep{noy1954motif} to select our group of motifs.

\cite{daranyi.2010.procamicus.1.29} has called attention to the need for research into the automation of extraction and annotation of motifs in folklore. Further work by \cite{daranyi.2012.analesdoc.15.x} has determined that motifs may not be the highest level of abstraction in narrative, \cite{daranyi.2012.proccmn.3.2} have made substantial headway towards using motifs as sequences of ``narrative DNA'', and \cite{ofek.2013.proccmn.4.166} have demonstrated learning tale types based on these sequences. \cite{declerck.2012.procdhc.x.x} have also done work on converting electronic representations of TMI and ATU to a format that enables multilingual, content-level indexing of folktale texts, building upon past work \citep{declerck2011linguistic}. Currently, this work appears to be focused on the descriptions of motifs and tale types, without reference to the stories. In this work, we take the first steps towards automatically detecting motifs in folklore as well as verifying this system with gold-standard data created by in-culture annotators.

With regard to analyzing motif annotation schemes, \cite{karsdorp2012search} present an analysis of the degree of abstraction present in the ATU catalog and the methods used to note what motifs belong to a given tale type. They find the ATU annotation insufficient for analyzing recurring motifs across types, in that it the ATU scheme fails to capture commonalities across closely related types while also failing to provide sufficient detail. We aim to rectify this in our own annotation by providing detailed descriptions of motifs and providing a much clearer, more formal definition drawn from the work of \cite{yarlott2016learning}.

Additionally, \cite{yarlott2016learning} have done work towards a more formal definition of motif as well as providing a general framework for a potential motif extraction and detection system. Our system continues this work by begining the development of the proposed framework.

\section{Future Work \& Contributions}
\label{sec:contributions}
Currently, we aim to find features to continue developing the motif detector with the aim of substantially outperforming the preliminary metaphor results. We are also implementing a neural baseline to compare to the result of the pipeline in both performance and generalizability. We plan to complete the annotation study and finalize the analysis to further demonstrate the importance of motifs.

The contributions of this short paper are as follows: we have shown preliminary work on the annotation of motifs alongside preliminary agreement measures that suggest humans with a strong background in the relevant cultural group can reliably distinguish motific usage of motifs. We have described, in detail, our pipeline architecture, currently under development. We have also presented preliminary results on the use of metaphoricity as a feature for distinguishing motific usages of motif surface forms, which serve to demonstrate the gap between metaphoric and motific language with respect to automatic detectors, as well as the need for a detector specifically designed for motifs.

% \newpage
 
\begin{acknowledgements}
\noindent
This work was supported by DARPA via SBIR Phase II Prime contract FA8650-19-C-6017. Any opinions, findings, and conclusions or recommendations expressed in this material are those of the author(s) and do not necessarily reflect the views of the DARPA. 
\end{acknowledgements} 

\vspace{-0.25in}

{\parindent -10pt\leftskip 10pt\noindent
\bibliographystyle{cogsysapa}
\bibliography{acs}

}

% Leave a blank line before the closing brace to ensure the final 
% reference has the proper indentation. 

\end{document}